\documentclass[envcountsect]{ifacconf}
\usepackage{algorithm}
\usepackage[nolist,nohyperlinks]{acronym}
\usepackage{xcolor}
\usepackage{amssymb}
\usepackage{amsmath}
\usepackage{caption}
\usepackage{optidef}
\usepackage{graphicx}
\usepackage{url}
\usepackage{subcaption}
\usepackage{caption}
\usepackage{blindtext}
\usepackage{algpseudocode}
\usepackage{cleveref}
\usepackage{graphicx}      
\usepackage{natbib}        

\acrodef{LMT}{linear model tree}
\acrodef{AI}{artificial intelligence}
\acrodef{CFE}{counterfactual explanation}
\acrodef{DL}{deep learning}
\acrodef{XAI}{explainable artificial intelligence}
\acrodef{DT}{decision tree}
\acrodef{LORE}{LOcal Rule-Based Explainer}
\acrodef{DRL}{deep RL}
\acrodef{ML}{machine learning}
\acrodef{LIME}{Local Interpretable Model-agnostic Explainer}
\acrodef{NN}{neural network}
\acrodef{RL}{reinforcement learning}
\acrodef{PPO}{Proximal Policy Optimization}

\begin{document}
\begin{frontmatter}

\title{ Real-Time Counterfactual Explanations For Robotic Systems With Multiple Continuous Outputs\thanksref{footnoteinfo}} 

\thanks[footnoteinfo]{This work was supported by the Research Council of Norway through the EXAIGON project, project number 304843.}

\author[First]{Vilde B. Gjærum} 
\author[Second]{Inga Strümke} 
\author[First]{Anastasios M. Lekkas}
\author[Third]{Timothy Miller}

\address[First]{Department of Engineering Cybernetics, Norwegian University of Science and Technology, Trondheim, Norway, (e-mail: vilde.gjarum, anastasios.lekkas ,@ntnu.no).}
\address[Second]{Department of Computer Science, Norwegian University of Science and Technology, Trondheim, Norway ,(e-mail: inga.strumke@ntnu.no)}

\address[Third]{School of Computing and Information Systems, University of Melbourne, Melbourne, Australia(e-mail: tmiller@unimelb.edu.aus)}

\begin{abstract}                
Although many machine learning methods, especially from the field of deep learning, have been instrumental in addressing challenges within robotic applications, we cannot take full advantage of such methods before these can provide performance and safety guarantees. The lack of trust that impedes the use of these methods mainly stems from a lack of human understanding of what exactly machine learning models have learned, and how robust their behaviour is. This is the problem the field of explainable artificial intelligence aims to solve. Based on insights from the social sciences, we know that humans prefer contrastive explanations, i.e.\ explanations answering the hypothetical question \emph{``what if?''}. In this paper, we show that linear model trees are capable of producing answers to such questions, so-called \emph{counterfactual explanations}, for robotic systems, including in the case of multiple, continuous inputs and outputs. We demonstrate the use of this method to produce counterfactual explanations for two robotic applications. Additionally, we explore the issue of infeasibility, which is of particular interest in systems governed by the laws of physics.
\end{abstract}

\begin{keyword}
Explicability and transparency in Cyber-physical and human systems, Reinforcement learning and deep learning in control, data-driven control, autonomous robotic systems, explainable artificial intelligence for robotics, counterfactual explanations for robotic systems
\end{keyword}

\end{frontmatter}

\section{Introduction}

\Ac{AI} has shown to be useful for robotics, control, and autonomous systems in several ways, for example by \ac{DL} boosting the performance of image processing and thus also giving robots better perception.\Ac{RL} is a subfield of \ac{ML} where the agent learns through exploring its environment, and through trial and error improving its strategy. One of the main benefits of using \ac{RL} is that we do not need to label every prediction as right or wrong but can rather encourage desired behaviour through a reward function describing the wanted outcome of the agent's behaviour. \Ac{DRL} is well-suited for problems where the system and/or the environment cannot be modelled accurately making it a good approach for many challenging control problems. There are numerous examples of \ac{RL}-agents successfully performing robotic tasks \citep{Remman21b,curved_MartinsenLekkas2018, Haarnoja2019, Lillicrap2016Continuous}.
When a model is so complex that humans no longer can understand how they make their decisions or predictions we call it a \textit{black-box model}. For \ac{AI}-methods of black-box nature, such as deep neural networks, to reach their full potential in applications with real-life risk associated we need to better understand their inner workings. By using \ac{XAI}-methods we can get increased trust, better knowledge, improved control, and the ability to justify decisions and predictions~\citep{Adadi2018}. The fields of robotics and control have some additional requirements for the explanations given by \ac{XAI}-methods, such as:
\begin{itemize}
    \item explanations to be used under live operations must be given in real-time,
    \item the methods must be able to handle large, continuous input and output spaces
    \item the methods must be able to handle large datasets.
\end{itemize}
As seen in \cite{Lover21CAMS}, these are requirements that some \ac{XAI}-methods struggle to meet. In \cite{GjaerumJMSE21} it is shown that \acp{LMT}, a \ac{DT} with linear functions in the leaf nodes, are well-suited for acting as a post-hoc, explanation method for a \ac{RL}-agent performing docking of a surface vessel. In \cite{GjaerumJMSE21}, the \ac{LMT} gave explanations in the form of feature attributions that states a direct mapping from the inputs to the outputs. Another popular form of explanation is \acp{CFE}. \Acp{CFE} answers the hypothetical question \textit{``but what if (the state was different)?"} or the more specific question \textit{``why did you do this instead of that?"}.
A commonly used example for showing the utility value of counterfactual explanations is the case of someone getting a loan application rejected, and the importance of giving an \textit{actionable} explanation is emphasized. For example, \textit{``if you had more savings your loan application would have been accepted} is an actionable explanation since this is something that can be changed, whereas \textit{``if you were 10 years younger your loan application would have been accepted} is not an actionable explanation. Similar examples can be found in other fields, such as job and university applications, predicting the risk of disease in the future, or disbursing government aid~\citep{Mothilal2020}. However, this notion of actionable explanations does not apply to the field of robotics as it does not make sense to change the input features(the state). Instead, the focus should be more on the \textit{feasibility} of the counterfactual explanations, in the sense of whether or not the counterfactual state and the counterfactual action are physically possible~\citep{Guidotti2022}. In~\cite{GjaerumJMSE21}, it is shown that \acp{LMT} are capable of giving explanations in the form of feature attributions in real time for robotic applications. \Acp{LMT} are decision trees where the constant prediction in the leaf nodes is replaced by a linear function. Since the branch node and its splitting conditions divide the input space into different regions, and each region has a corresponding linear function, the \acp{LMT} constitute a piece-wise linear function. If the \Ac{LMT} has accurately enough approximated black-box model, such as a \Ac{NN}, it can be used as a post-hoc explanation method. In \cite{Carreira2021}, an exact method for computing \acp{CFE} for classification trees with both univariate and multivariate splits in the branch nodes is proposed. Regression trees with univariate splitting conditions are solved directly by exploiting the fact that the problem is separable within the leaf nodes, meaning that solving the problem in every leaf node and then choosing the leaf node with the best \ac{CFE} solves the problem globally. For the classification trees with multivariate splits, the problem of finding \acp{CFE} are expressed as a mixed-integer problem and the solution is found by using linear or quadratic problem solvers depending on whether the objective function is expressed as a linear or quadratic problem. A model-agnostic method for finding \acp{CFE} is presented in \cite{Karimi2019}. The problem of finding \acp{CFE} is expressed as a sequence of \textit{satisfiability}(SAT) problem. Neither \cite{Carreira2021} nor \cite{Karimi2019} are applicable to regression problems with multiple outputs and thus not applicable to most robotic applications. In \cite{Sokol19}, \acp{CFE} are found using a \textit{leaf-to-leaf counterfactual distance matrix} describing how much an instance belonging to a leaf node would have to change to belong in another leaf node. However, the method proposed is for classification trees and details for implementation lacks. Both \ac{LORE}~\cite{Guidotti2019} and FOILTREE~\cite{venderwaa2018} build a \ac{DT} locally around the instance to be explained and searches through the \ac{DT} for \acp{CFE} but neither apply to robotic applications since they do not apply to regression problems with multiple outputs. Another problem with methods that builds a new local interpretable model for every explanation is that they are often not fast enough to be used in real-time \citep{Lover21CAMS}. In this paper, we present an algorithm for finding counterfactual explanations from an \ac{LMT} and show that this method is applicable to robotic applications.

The paper's main contributions are as follows:
\begin{itemize}
    \item Algorithm for getting counterfactual explanations in real-time for a black-box model through an \ac{LMT} serving as a surrogate model.
    \item Using the algorithm on two robotic applications, one with singular, continuous outputs and one with multiple, continuous outputs.
    \item Identify the issue of infeasibility that arises when producing \acp{CFE} for real-world problems where the laws of physics apply.
    \item Suggesting remedies for avoiding infeasible counterfactuals.
\end{itemize}

The paper is structured as follows. In \Cref{sec:LMTs}, \Acp{LMT} will be presented, followed by an introduction to \acp{CFE} in \Cref{sec:CFEs}. In \Cref{sec:Methods}, we will present how \acp{LMT} can be used to find \acp{CFE}. In \Cref{sec:results}, the results are presented and the discussion is given in \Cref{sec:discussion}.
\section{Background}
In this section, the necessary background will be introduced. First, the \acp{LMT} will be introduced in \Cref{sec:LMTs} followed by the two applications we test the \acp{LMT} on in \Cref{sec:test_apps}. Finally, the \acp{CFE} will be introduced in \Cref{sec:CFEs}.
\subsection{Linear Model Trees}\label{sec:LMTs}

\Acp{DT} consist of branch nodes and leaf nodes. The branch nodes have a univariate splitting condition, meaning they split the data based on whether or not a certain input feature is smaller or bigger than a threshold. In this way, the branch nodes split the input space into distinct regions and each region has a corresponding leaf node. \Acp{LMT} are decision trees with linear functions in the leaf nodes and \acp{LMT} are piece-wise linear function approximators. As presented in \cite{GjaerumNeurocomputing21}, several methods for building \acp{LMT} for robotic applications exist but for this work \acp{LMT} trained with the method presented in \cite{GjaerumJMSE21} will be used.

\subsection{Test applications}\label{sec:test_apps}
In this section, we introduce the two robotic applications for which the \Ac{LMT} produced counterfactual explanations.

\subsubsection{Pendulum:}
The inverted pendulum\footnote{See \url{https://www.gymlibrary.ml/environments/classic_control/pendulum/}{}} is a classic control theory problem where the goal is to balance the pendulum in the upright position by applying force to the free end of the pendulum. The pendulum has three states, namely the angular velocity $\Dot{\theta}$ and the coordinates of the free end of the pendulum $x$ and $y$. The force applied to the free end of the pendulum is continuous and is the only action for this environment. A \ac{NN} has been trained with the \ac{RL}-method \ac{PPO}~\citep{schulman2017} to balance the pendulum.

\subsubsection{Docking environment:}
The docking agent and the harbour environment are thoroughly presented in \cite{GjaerumJMSE21}. The vessel has eight input features describing the vessel's relative position and velocity in the harbour. The vessel has three thrusters, one tunnel thruster at the front and two azimuth thrusters at the back. The tunnel thruster is controlled by setting the force of the thruster, while the azimuth thrusters are controlled by setting the force and angle of the thruster. The \ac{NN} that performs the docking was trained by \ac{PPO} as originally presented in \cite{Rorvik2020}.

\subsection{Counterfactual explanations}\label{sec:CFEs}

\begin{defn}[Counterfactual explanation  \citep{Guidotti2022}]
\label{def:CFE}
Given a classifier $b$ that outputs the decision $y = b(x)$ for an instance $x$, 
a counterfactual explanation consists of an instance $x'$ such that the decision for $b$ on $x'$ is different from $y$, 
i.e., $b(x') \neq y$, and such that the difference between $x$ and $x'$ is minimal.
\end{defn}

Following this definition, a \ac{CFE} can be formulated as: \\
If state \textit{s} had been $\Delta s$ different, the corresponding action \textit{a} would have been $\Delta a$ different.

The most frequently asked contrastive question that applies to robotic applications is \textit{``Why is action A used in state S, rather than action B?"} \citep{Krarup21}.
The classifier $b$ in Definition~\ref{def:CFE} is usually a black-box model such as a deep neural network, and we are looking for another instance with as similar input features as our original instance but with a different output. The meaning of a different output is straightforward for classification problems as the counterfactual example will be the closest instance (based on some distance metric on the input features) with a different class as output. This is more challenging for regression problems since it is not as clear what the counterfactual action is since the meaning of \textit{different enough output} is context-dependent. Counterfactual explanations for problems with multiple, continuous input and output features can be defined as follows.

\begin{defn}[Cont. counterfactual explanation]
\label{def:CFE_cont}
Given a predictor $b$ that outputs a continious prediction of dimension $n$ $y = b(x)$ for a continuous instance $x$ of dimension $m$, 
a counterfactual explanation consists of an instance $x'$ such that the distance from $b$'s prediction on $x'$ from $y$ is maximal, 
i.e., $b(x') \neq y$, and such that the difference between $x$ and $x'$ is minimal.
\end{defn}

\section{Method}\label{sec:Methods}

\Acp{CFE} can be found by using an \ac{LMT} as shown in Alg. \Cref{alg:CFE_from_LMTS} by ordering the leaf nodes from closest to furthest and then solving an optimization problem within that region instead of over the entire state space. 
\begin{algorithm}[tb]
\caption{CFEs from LMTs}\label{alg:CFE_from_LMTS}
\begin{algorithmic}
\Require 
\\ $x$: instance to be explained
\\ \textit{num\_exp}: number of explanations  wanted
\\ \textit{constraints}: known constraints for input and output 
\\ $ordered\_leaf\_nodes$: the trees leaf nodes ordered from closest to furthest relative to the leaf node $x$ belongs to 
\\ $f$: Objective function \\

\State i = 0
\While{i $<$ \textit{num\_exp}}:
    \State $leaf\_node \gets ordered\_leaf\_nodes[i] $
    \State $constraints \gets $boundaries of $leaf\_node$
    \State $x' \gets $ minimize $f$ subject to $constraints$
    \State $y' \gets $ black\_box($x'$)
    \State 
    
\EndWhile
\end{algorithmic}
\end{algorithm}
To ensure the \textit{correctness} of the explanations, the \ac{LMT} is only used to locate the counterfactual example and the black-box is used when formulating the explanation.

We make the following two main assumptions:
\begin{enumerate}
    \item We assume that the \ac{LMT} is successfully trained and thus has approximated the black-box model with sufficient accuracy \citep{GjaerumNeurocomputing21}.
    \item We assume that the tree has placed leaf nodes of regions that are similar in vicinity of each other.
\end{enumerate}

\subsection{Leaf node ordering}
\begin{figure}
    \centering
    \includegraphics[width=0.35\textwidth]{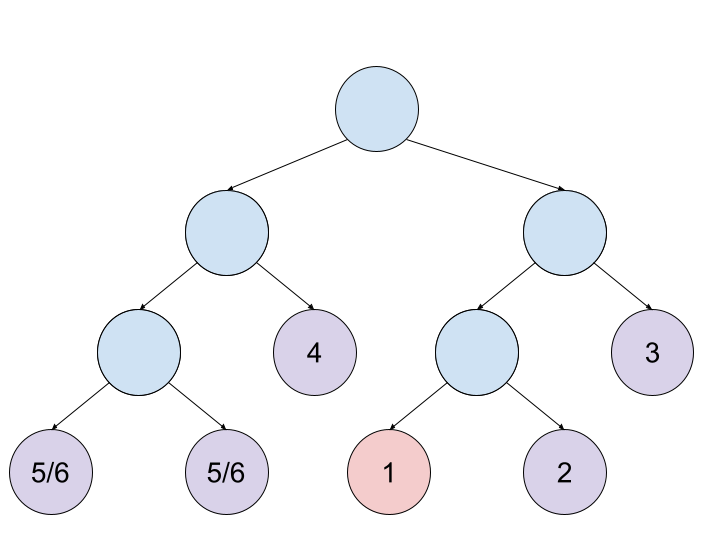}
    \caption{Illustration of how the leaf nodes will ordered if the instance to be explained belongs to the red leaf node.}
    \label{fig:leaf_node_ordering}
\end{figure}

To avoid searching through the entire state space for counterfactual explanations, we exploit the fact that the \ac{LMT} has already split the state space into regions. We want to order the regions by how close they are to the instance to be explained. Since each region corresponds to a specific leaf node we can do this by ordering the leaf nodes, which again can be done by looking at the structure of the tree. This ordering of leaf nodes can be found by counting how many branch nodes must be traversed to get from the instance to every other leaf node. An example of how the leaf nodes would be ordered is shown in \Cref{fig:leaf_node_ordering} and the algorithm for ordering the leaf nodes is presented in \Cref{alg:LNO}.

\begin{algorithm}[tb]
\caption{Leaf node ordering}\label{alg:LNO}
\begin{algorithmic}
\Require instance $x$ to be explained
    \State $\mathcal{L}_c \gets$ the leaf node x belongs to \Comment{$\mathcal{L}_c$ is the current leaf node}
\State $LNO \gets \mathcal{L}_c$ \Comment{List containing the \textit{L}eaf \textit{N}ode \textit{O}rder}
\State $\mathcal{L}_c \gets Parent(\mathcal{L}_c)$
\While{$\mathcal{L}_c$ is not the root node}
    
    \State Traverse the subtree starting for $\mathcal{L}_c$ 
    \State Add leaf nodes to $LNO$ in the order they're found.
    \State $\mathcal{L}_c \gets Parent(\mathcal{L}_c) $ 

\EndWhile
\end{algorithmic}
\end{algorithm}

\subsection{Counterfactual explanations from optimization}
Given that we already have the leaf nodes ordered from closest to furthest relative to the leaf node the instance to be explained $x$ belongs to, the counterfactual example can be found by solving the optimization problem given in \ref{eq:opt_problem}.

\begin{equation}\label{eq:opt_problem}
\begin{matrix}
\min_{x'} &  z = \mid x-x'\mid - (y-y')^2 - sparsity(x') \\\vspace{5pt}
\textrm{s.t.} & x'_{j_i} \leq t_i,& i \in \mathcal{P}_{left} \\\vspace{5pt}
&  x'_{j_i} > t_i,& i \in \mathcal{P}_{right}\\ \vspace{5pt}
& x'_{j_i} > t_i,& i \in \mathcal{B}_{upper}^{input} \\\vspace{5pt}
& x'_{j_i} < t_i,& i \in \mathcal{B}_{lower}^{input} \\\vspace{5pt}
& y'_{j_i} > t_i,& i \in \mathcal{B}_{upper}^{output} \\\vspace{5pt}
& y'_{j_i} > t_i,& i \in \mathcal{B}_{lower}^{output} \\\vspace{5pt}
& y' = f_l(x') &
\end{matrix}
\end{equation}

where $x'$ is the counterfactual example and its corresponding output $y'$. The left and right parent nodes along the path from the root node to the leaf node are denoted by $\mathcal{P}_{left/right}$, while $t_i$ is the threshold used in the splitting condition on input feature $j$. The lower and upper boundaries on the input are given by $\mathcal{B}^{input}_{lower/upper}$, while the lower and upper boundaries on the output are given by $\mathcal{B}^{output}_{lower/upper}$. The objective function is given by $z$ in \ref{eq:opt_problem} and the expression consists of three parts:
\begin{itemize}
    \item $|x - x'|$: The distance between the input values of the instance to be explained and the counterfactual example should be minimized
    \item $(Y - y')^2$: The distance between the output value of the instance to be explained and the counterfactual example should be maximized
    \item sparsity($x'$) and sparsity($y'$): Simple explanations are preferred and thus it is better with explanations where a few features have been changed a lot rather than many features changed slightly.
\end{itemize}
For different applications, it may be beneficial to change these metrics with for example different distance functions or different sparsity measurements.

\subsubsection{Requesting specific explanations}
Given the objective function in \ref{eq:opt_problem}, we are searching for the counterfactual problem that balances finding an example with a as small as possible change in input features while having a as big as possible change in the output feature. However, if a specific explanation is requested, such as \textit{``Why was the action y taken instead of action Y?"}, we are no longer searching for a counterfactual example in general but rather the counterfactual example with the smallest change in the input and output as close as possible to $Y$. This specific counterfactual can be found by using
\begin{equation}\label{eq:req_expl}
     z = |x-x'| - (Y-y')^2 - sparsity(x') - sparsity(y'),
\end{equation}
where 
\begin{equation}
    sparsity(x') = | x - x' |_0 ,
\end{equation}
and 
\begin{equation}
    sparsity(y') = | y - y' |_0 .
\end{equation}
\subsection{Pipeline}

\begin{figure}
    \centering
    \includegraphics[width=0.45\textwidth]{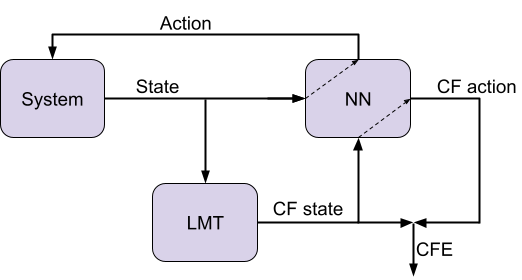}
    \caption{Pipeline describing how the system, the \ac{NN}, and the \ac{LMT} works together.}
    \label{fig:pipeline}
\end{figure}

How the system, the \ac{NN}, and the \ac{LMT} are connected is illustrated in \Cref{fig:pipeline}. For both the case of the pendulum and the docking agent the controller is a \ac{NN} but since the \ac{LMT} is a model-agnostic method any type of model can be used to control the system. The \ac{NN} gets the state from the system and returns the action for that state. The \ac{LMT} receives the same state for the system and calculates what the counterfactual state is, gives this counterfactual state to the \ac{NN} and the \ac{NN} returns the counterfactual action. By combining the counterfactual state and action we get the counterfactual explanation. By doing it this way we are certain that the counterfactual explanation is true because we are sure that this is the action the \ac{NN} would have taken given the state found by the \ac{LMT}.

\section{Infeasible explanations}
Given a black-box controlling a robot operating in the real world, we argue that explanations must make sense from a physical point of view. However, this may contradict the desire for a sparse explanation. Take the inverted pendulum as an example. Following the desire for sparse explanations, a counterfactual example where only one of $x$ or $y$ is changed is preferred. However, since the free end of the pendulum always will be at a point along a circle with a radius the length of the pendulum, thus, with only one exception, all counterfactual examples changing only one of the two coordinates will be physically impossible and thus make no sense. We address this through feature engineering as presented in Sec. \Cref{sec:feature_engineering}, adding constraints to the optimization problem as presented in \Cref{adding_constraints}.

\subsection{Feature engineering}\label{sec:feature_engineering}
One way of ensuring that the tree does not return infeasible counterfactuals is by making sure that the features, both input and output, are not dependent. This can be achieved using feature engineering to find new features that are independent but represent the original problem (or an approximation of it).

In the case of the pendulum problem, this could be done by using $\theta$ and $\Dot{\theta}$ as input features instead of $x$,$y$, and $\Dot{\theta}$.

\subsection{Adding constraints}\label{adding_constraints}
If the relationship between the dependent features is known and can be formalized as a function they can be added to the optimization solver. It is important to note that the constraint functions form affects which optimization solvers can be used. For the case of the pendulum, the following constraint can be added:
\begin{equation}
    L = x^2 + y^2,
\end{equation}
where $L$ is the length of the pendulum, and $x$ and $y$ the position of the end of the pendulum.

\section{Results}\label{sec:results}



In this section, we show that the \acp{LMT} are capable of giving counterfactual explanations for robotic systems with both singular and multiple continuous inputs and outputs in real time. For both the inverted pendulum environment and the docking environment a \ac{NN} is trained to perform the respective tasks, and an \Ac{LMT} is built to approximate the \Acp{NN}. The \ac{LMT} approximating the docking agent was trained and presented in \citep{GjaerumJMSE21}, while the \ac{LMT} and the \ac{NN} for the pendulum were built for this paper. In \Cref{tab:CFE_times}, the average time it took to compute the counterfactual explanations for 250 different states on an Intel \textregistered Core\texttrademark i9-9980HK CPU @ 2.40GHz. The explanations can be computed within a quarter of a second which is faster than humans can interpret the explanations and the explanations are thus suitable for use in real-time. The \ac{LMT} giving explanations for the pendulum-agent has 220 leaf nodes, while the \ac{LMT} giving explanations fo r the docking agent has 312 leaf nodes. The most time consuming part of the algorithm is the ordering of all the leaf nodes. This can be speeded up by only ordering the a limited number of the closest leaf nodes instead of all of the leaf nodes. This is especially helpful for large trees.
\begin{table}[tb]
    \centering
    \captionsetup{width=6cm}
    
    \begin{tabular}{cc}
        \textbf{ Application} &  \textbf{Average time} \\ \hline \hline
         Pendulum & 0.21 s \\ \hline
         Docking env. & 0.036 s\\ \hline
    \end{tabular}
    \caption{Average time for computing the counterfactual explanation for each of the test applications.}
    \label{tab:CFE_times}
\end{table}

\begin{figure}
    \centering
    \begin{subfigure}{0.20\textwidth}
        \includegraphics[width=0.75\linewidth]{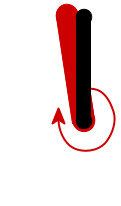}
        \caption{\label{fig:pendulum_expl1}}
    \end{subfigure}\\
    \begin{subfigure}{0.20\textwidth}
        \includegraphics[width=0.75\linewidth]{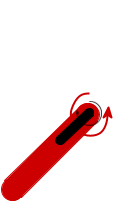}
        \caption{\label{fig:pendulum_infeasible_expl}}
        \end{subfigure}
    \begin{subfigure}{0.20\textwidth}
    \includegraphics[width=0.75\linewidth]{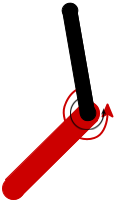}
    \caption{\label{fig:pendulum_fe_expl}}
    \end{subfigure}
    \caption{The pendulums state is given by the red rod and the torque applied by the black-box model is shown in red. The counterfactual explanation is given by the counterfactual state (the black rod) and the counterfactual action (black torque).\\
    (\protect\subref{fig:pendulum_expl1}) shows a feasible counterfactual explanation,\\
    (\protect\subref{fig:pendulum_infeasible_expl}) shows an infeasible counterfactual explanation,\\ and\\
    (\protect\subref{fig:pendulum_fe_expl}) shows the corresponding feasible counterfactual explanation found by using feature engineering.
    } 
\end{figure}

In~\Cref{fig:pendulum_expl1}, a counterfactual explanation for the inverted pendulum is shown. The thick, red line is the position of the pendulum, while the black line is the position of the pendulum in the counterfactual state. The torque applied to the pendulum by the \ac{NN} is shown with a red, circular line while the counterfactual action is shown with a black, circular line.


In~\Cref{fig:pendulum_infeasible_expl}, an example of an infeasible counterfactual explanation is shown. This explanation is infeasible because the counterfactual state is infeasible. In fact, when building an \ac{LMT} using the features $\theta$ and $\Dot{\theta}$ instead of $x$,$y$, and $\Dot{\theta}$, as discussed in \Cref{sec:feature_engineering}, we find that the \Ac{LMT} always gives feasible counterfactuals.This can be seen in \Cref{fig:pendulum_fe_expl}, where a feasible counterfactual explanation for the same state as in \Cref{fig:pendulum_infeasible_expl} is shown.

\begin{figure}[tb]
    \centering
    \includegraphics[width=\linewidth]{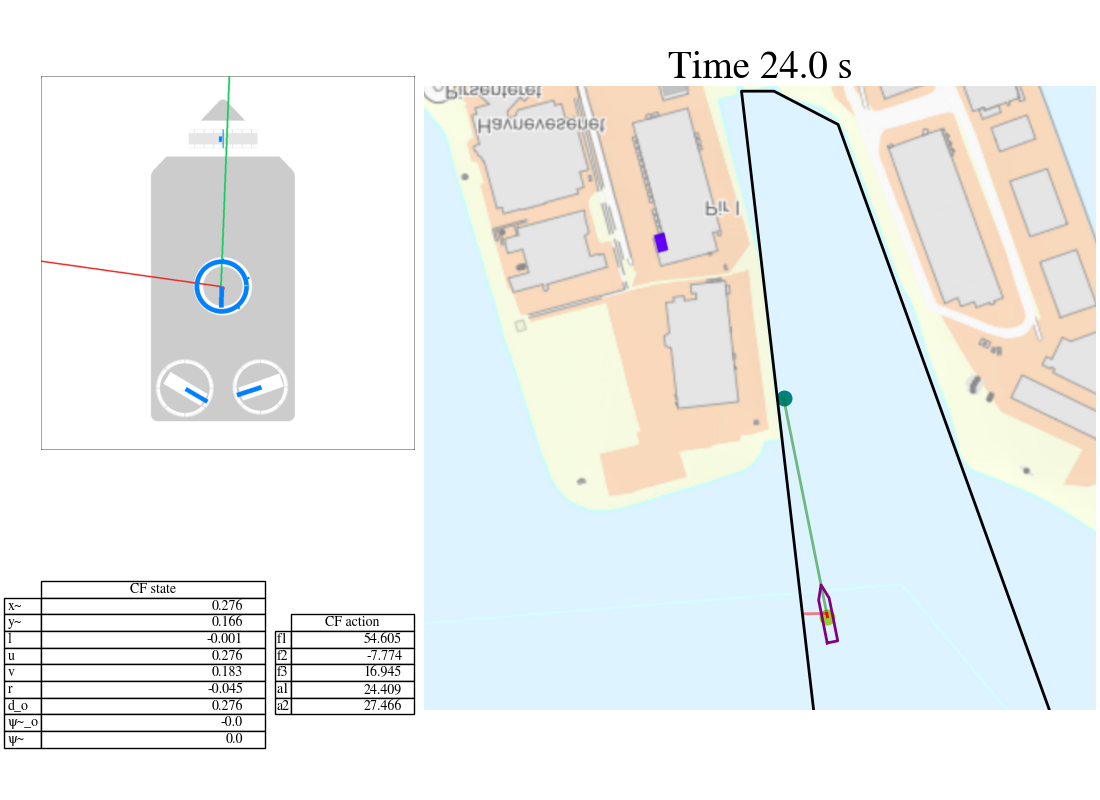}
    \caption{The table shows the counterfactual explanation by stating how much the state and the action would change in the counterfactual example for the situation given by the vessels point of view(top left) and how the vessel is situated in the harbour (right).}
    \label{fig:docking_expl}
\end{figure}

In \Cref{fig:docking_expl}, a counterfactual explanation for the docking agent is shown. Since the vessel has so many input and output features we found it easier to express the counterfactual explanation in a table rather than in words. Still, the explanation could be communicated faster with appropriate visualizations.

\section{Discussion}\label{sec:discussion}

We have shown that \acp{LMT} are suitable for generating counterfactual explanations even for complex robotic systems with multiple, continuous outputs. Which counterfactual explanation is found is determined solely by \ref{eq:opt_problem}, and the tuning of this function is crucial because it determines the trade-off between a large distance in the output, a small distance in the input and the sparsity of the explanation. Since the \ac{LMT}  only identifies the counterfactual state, while the \ac{NN} is used to complete the counterfactual explanation by finding the counterfactual action, we know that the counterfactual explanation is necessarily true. However, we do not know whether there exist better (in terms of distance in input and output) counterfactual explanations that could have been found by using another cost function.
The \acp{CFE} could also say something about the agents or the environments stability around a certain point by looking at how far (in the input space) we have to look before finding a significantly different output. Actions that are changing a lot, even when the state is just slightly different, can be due to either the agent being in an especially complex region or that the agents behaviour is unstable.
As shown for the pendulum case, the \ac{LMT} may find infeasible states if the input features are not independent. One way of handling this problem is by performing feature engineering on the input features so that they become independent. 
This approach was successful in the case of the pendulum but would be significantly harder in a more complex environment, as is the case for the docking agent. In some cases, not all the relevant dependencies are defined or even known. If they are known, they can be added to the optimization problem as constraints, which again can make the optimization problem harder to solve. Evaluating the usefulness of an explanation is difficult because explanations are subjective, and different recipients prefer different types of explanations. Additionally, how an explanation is communicated is sometimes as important as the explanation itself, especially for complex systems. Therefore, future work should include investigating both to what extent these explanations can improve the understanding of the system and how to communicate these explanations most effectively. The usefulness of infeasible explanations should also be investigated.


\bibliography{ifacconf}             
                                                   







\end{document}